# BPLF: A Bi-Parallel Linear Flow Model for Facial Expression Generation from Emotion Set Images

Gao Xu[1]*, Yuanpeng Long[2]*, Siwei Liu[1], Lijia Yang[1], Shimei Xu[3], Xiaoming Yao[1,3], Kunxian Shu[1#]

1. School of Computer Science and Technology, Chongqing Key Laboratory on Big Data for Bio Intelligence, Chongqing University of Posts and Telecommunications, Chongqing 400065 China
2. School of Economic Information Engineering, Southwestern University of Finance and Economics, Chengdu 611130, China
3. 51yunjian.com, Hetie International Square, Building 2 Room404, Chengdu, Sichuan, China

**\* Equal contribution**
**# Corresponding author:** Kunxian Shu (e-mail: shukx@cqupt.edu.cn).

# Abstract

The flow-based generative model is a deep learning generative model, which obtains the ability to generate data by explicitly learning the data distribution. Theoretically its ability to restore data is stronger than other generative models. However, its implementation has many limitations, including limited model design, too many model parameters and tedious calculation. In this paper, a bi-parallel linear flow model for facial emotion generation from emotion set images is constructed, and a series of improvements have been made in terms of the expression ability of the model and the convergence speed in training. The model is mainly composed of several coupling layers superimposed to form a multi-scale structure, in which each coupling layer contains 1*1 reversible convolution and linear operation modules. Furthermore, this paper sorted out the current public data set of facial emotion images, made a new emotion data, and verified the model through this data set. The experimental results show that, under the traditional convolutional neural network, the 3-layer 3*3 convolution kernel is more conducive to extracte the features of the face images. The introduction of principal component decomposition can improve the convergence speed of the model.

**Keywords**: facial emotion generation; generative model; bi-parallel linear flow model; deep learning

# 1 Introduction

The LeNet-5 proposed by LeCun[1] et al. has been successfully applied to handwritten digit recognition, providing theoretical basis for deep learning. Deep learning research methods include two major categories: supervised learning and unsupervised learning. At present, most ideas adopt the method of supervised learning, using known data to build a model to predict unknown data, and corresponding data sets are needed as support. Since the collection of data sets and the labeling of data categories need a lot of tedious work , research on training deep models in semi-supervised or unsupervised ways has attracted attention. Lee[2] et al. used the unlabeled samples predicted by the current model, and then used them to train the model. Sajjadi[3] established unsupervised regularization terms to train classifiers of multiple categories to obtain the decision boundary. Rasmus[4] et al. used the sum of supervision and denoising losses as the loss function, and combined unsupervised and supervised learning. Dinh[5] et al. obtained inspiration from VGG and proved that the L2 paradigm is more effective in expression recognition networks. In recent years, the generative adversarial networks[6] have been widely used in image generation and style transfer generating many interesting applications [7][8].

At present, in the field of facial emotion research, it is common to construct discriminant models[9][10], which directly learn decision functions or probability distributions. The basic idea is to establish a discriminant function, regardless of the generation model generated by the sample, and to directly study the prediction model . The discriminant model still has certain limitations when dealing with actual problems. It fits the decision function y=f(x) or the conditional probability distribution p(y|x), only reflecting heterogeneous data. The difference between the two cannot reflect the characteristics of the training data itself, and thus is a simplification of the learning problem[11]. The generative model learns the joint probability density distribution y=f(x,y), which can be learned from statistics. The angle represents the distribution of data, which can reflect the similarity of similar data itself.

The widely used generative models at this stage use either optimization upper bounds[12][13][14], or countermeasures[6][15][16] to avoid probability calculations, in order to approximate the true distribution. The generative models can achieve the suitable fitting to the data distribution. Compared with these generative models, the flow model is an optimal model which can directly calculate the integral formula that transforms the distribution. As such, it is one of the biggest advantages of the flow model[17].

# 2 Background information

## 2.1 Research on facial emotions

Darwin[18] believes that human emotions are innate and universal. In the 1960s, psychologists confirmed the universal nature of emotions through scientific research. Furthermore, after analysis and summary, Ekman[19] and others put forward six basic human expressions, namely: anger, disgust, fear, happiness, sadness and surprise. There are also some psychologists who believe that emotions are influenced by culture and gradually formed with life experiences, which impacts in the inference of emotional experience and the intensity analysis of facial expressions.

Ekman and Paual, etc.[20] started with facial tissue muscles study and conducted long-term research on the control of facial expressions. They roughly divided the face muscles into 46 independent motion units, which are described as the controlled area and the corresponding expressions. Many images are used to exemplify this. This is the facial motion unit analysis method. The advantage of this method lies in its intuitive description of facial movement changes. At the same time, due to the complex facial muscles, it is a labor-intensive work.The labeling work is time-consuming and the system encoding runs slowly. The emergence of the MPEG-4 facial motion parameter method has solved this problem[21], which is based on facial muscles. Movement presents a complete collection of facial movements. The method adopted is to create a public face, and then visualize the details according to different faces. The motion parameters of the face are compared with the motion parameters of the neutral face, which greatly improves the recognition rate of facial expressions and reduces a lot of repetition work.

Ying-li Tian[22] proposed that the Cabor wavelet's recognition of facial units which attracted attention. The idea is similar to motion coding. The face is divided into upper and lower parts and marked as independent motion units. Geometric features are introduced for face recognition. In addition, research on human faces usually focuses on face detection. There are corresponding algorithm implementations in traditional methods and deep neural networks. The representative traditional method is Harr-AdaBoost[23] model. The representative model based on the deep neural network is MT-CNN[24].

Harr-AdaBoost connects multiple strong classifiers together. The strong classifier is composed of multiple weighted weak classifiers. Based on AdaBoost, it uses Haar-like wavelet features and integral maps for detection. Viola and Jones[25] designed more effective features for face detection and trained a strong classifier for cascading. Rainer Lienhart and Jochen Maydt[26] extended this work, and formed the current Haar classification in OpenCV.

MT-CNN constructs an image pyramid for the input images to adapt to different sizes of faces, making the model a multi-scale structure. This model contains three networks. After P-Net is processed, the face candidate frame is obtained. The Fully Convolution Network is used to obtain the coordinates of the boundary, and Non-Maximum Suppression is contained. Boundary regression is used to correct the face frame. The role of R-Net is to further eliminate redundant non-target candidate frames, filter and merge the results. Then O-Net is used to filter face candidate frames and detect face key points.

## 2.2 Flow model with reversible structure

The framework of the flow model is composed of a stack of multiple coupling layers. In the previous work, the coupling layer has undergone a series of evolution and optimization. In NICE[27], the concept of "additive coupling layer" is proposed, which divides the D dimension input data x into two parts according to a certain division: $x_1$, $x_2$, respectively carry out the following transformations:

$$\begin{cases} h_1 = x_1 \\ h_2 = x_2 + m(x_1) \end{cases} \quad (2\text{-}1)$$

Such a linear transformation is reversible, where m is an arbitrarily complex function of $x_1$. It will not affect the reversibility of the linear transformation, and is implemented as multiple fully connected layers (5 hidden layers, each hidden layer uses 1000 nodes, using Relu activation function). The dimensions of $h_1$ and $h_2$ are the same as those of $x_1$ and $x_2$, followed by $h_1$, $h_2$. These parameters will be used as input again to enter the next coupling layer. The above change is an identity transformation, that is, a single transformation cannot achieve relatively strong nonlinearity. Therefore, multiple such simple transformations are required to be stacked to increase the expressive ability of the model:

$$x = h^{(1)} \leftrightarrow h^{(2)} \leftrightarrow h^{(3)} \leftrightarrow \ldots \leftrightarrow h^{(n-1)} \leftrightarrow h^{(n)} = z \qquad (2\text{-}2)$$

In the above transformation process, in order to achieve an significant transformation, the order of each dimension needs to be disrupted. For example, in NICE, it is implemented through simple cross-coupling.

$$\begin{cases} h_1 = x_1 \\ h_2 = s(x_1) \otimes x_2 + t(x_1) \end{cases} \qquad (2\text{-}3)$$

s and t are arbitrary functions of $x_1$, and the symbol $\otimes$ means Hadamard product. Compared with the interleaving method used in NICE to stack the coupling layers, RealNVP[28] uses a random method to scramble the two-dimensional vectors, which can make the information exchange more efficient.

Furthermore, in the Glow[29] model, the RealNVP model is simplified and standardized. Its biggest contribution is the introduction of 1*1 convolution instead of the previous operation of disrupting the channel. Here, the 1*1 convolution is the evolution of the generalization of the permutation operation, and the permutation matrix is replaced by a trainable parameter matrix, as shown in the formula below:

$$h = xW \qquad (2\text{-}4)$$

At the same time, in order to ensure the reversibility of W, it is usually initialized as a random orthogonal matrix during training. Glow demonstrated the powerful generation ability of the flow model. The precise latent variables showed a good realization effect on the face attributes, which has attracted widespread attention. For example, f-VAEs[17] is produced by combining the advantages of the variational auto encoder. The flow-based model is used to solve the problem of blurring image generation.

# 3. Methodology

## 3.1 Calculating Jacobian matrix

Suppose a function F maps $R^n$ to $R^m$, and this function is composed of m real functions.

$$\begin{cases} y_1(x_1,x_2...x_n) \\ ... \\ y_n(x_1,x_2...x_n) \end{cases} \qquad (3\text{-}1)$$

Then the matrix formed by the partial derivatives of these functions is the Jacobian matrix:

$$J_F(x_1,...,x_n) = \begin{bmatrix} \frac{\partial y_1}{\partial x_1} & ... & \frac{\partial y_1}{\partial x_n} \\ \vdots & \ddots & \vdots \\ \frac{\partial y_m}{\partial x_1} & ... & \frac{\partial y_m}{\partial x_n} \end{bmatrix} \qquad (3\text{-}2)$$

As mentioned above, the essence of the flow model is to execute calculation and construct a bijection from the known distribution to the unknown distribution, so as to achieve the generation of samples of the target distribution. According to the variable transformation, the marginal log likelihood of each random sample can be written:

$$\log p_\theta(x) = \log p_\theta(z) + \sum_{i=1}^{L} \log|\det(\frac{dh_i}{dh_{i-1}})| \qquad (3\text{-}3)$$

L represents the number of coupling layers, and $h_i = f_i^{-1}(h_{i-1})$ is the number of coupling layers. The output is initialized to $h_0 = x$.

It can be seen from this, to achieve $z = f(x)$, creating an easy-to-calculate Jacobian matrix is needed. This matrix is usually designed as a diagonal matrix when designing a flow-based model. This is indeed the case for the affine coupling layer in Glow.

$$\begin{bmatrix} I & 0 \\ A & \text{Diagonal}(\beta_{d+1}, \beta_{d+2}, ..., \beta_D) \end{bmatrix} \qquad (3\text{-}4)$$

Among them, D represents the dimension of the data, and d represents the dimension of the first part of the identity transformation. It can be seen that the value of its determinant is the sum of the logarithms of the diagonal elements of the Jacobian matrix:

$$\log|\det(\frac{dh_i}{dh_{i-1}})| = \sum(\log|\text{diag}(\frac{dh_i}{dh_{i-1}})|) \qquad (3\text{-}5)$$

## 3.2 Bi-Parallel Linear Flow

We are inspired by the dynamic linear transformation and the grouping idea in the Capsule network. Here, we proposes a linear transformation with stronger parallelism:

$$\begin{cases} y_i^1 = s_1 \odot x_i^1 + \mu_1 \\ y_i^2 = s_2 \odot x_i^2 + \mu_2 \end{cases} \quad (3\text{-}6)$$

For the quuation $s_j, \mu_j = g_{j-1}(x_{j-1})$, $x_0, z_0$ is initialized to the default value. The similar calculation structure between linear transformations and the dependence between them enable the model to have a stronger expressive ability while taking into account the local correlation of the images. Unlike autoregressive models or sequence models, such dependence is included. It only exists in a single coupling structure, so there is no long-range dependency and will not affect the parallel computing capability of the model. Correspondingly, the inverse transformation of the bi-parallel linear flow is as follows:

$$\begin{cases} x_i^1 = (y_i^1 - \mu_1)/s_1 \\ x_i^2 = (y_i^2 - \mu_2)/s_2 \end{cases} \quad (3\text{-}7)$$

Based on the above research, this paper proposes a new flow-based parallel linear calculation module as shown in figure 1.

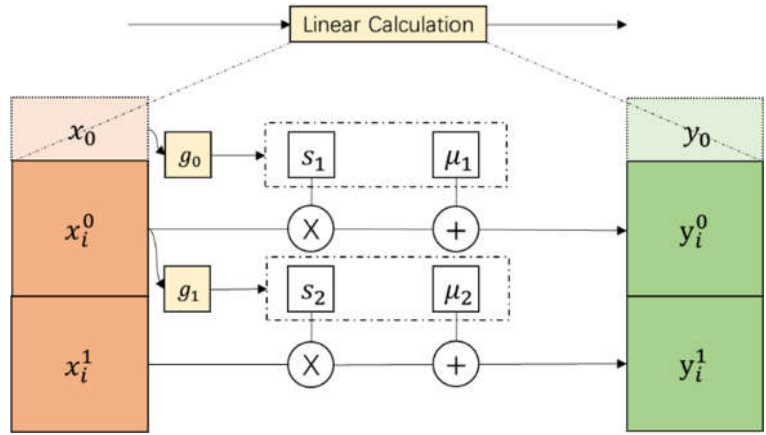

Figure 1 Linear calculation module in BPLF. Here g can be an arbitrarily complex function, usually implemented in a deep neural network.

Because the facial emotion image data is high-dimensional, the mapping ability of a single linear computing module is limited. Based on the above two improvements, this paper proposes a parallel linear flow PBLF. Inspired by Glow, the parallel linear flow is formed by stacking multiple identical linear calculation modules. Each calculation module includes 1*1 reversible convolution and linear calculation. The Decode of the model is defined as the process of generating the latent variable Z from the training data X. The model is composed of "D" modules. The function of each module is to perform calculations and divide the output into two parts. One part is passed to the next module, and the other part is directly used as output. The model as a whole constitutes multiple parts and scale structure. Before the data enters the calculation, singular value decomposition is performed on the basis of Squeeze. The data X is the operation of truncating the singular value of the image before the Squeeze operation, that is, the largest part of the singular value is retained for reconstruction. Then Squeeze is performed. From the sampling, Z in the known distribution. The process of performing the above model in reverse order is defined as the Encode of the model. Since the structure of parallel linear transformation is similar to that of inverse transformation, only the overall structure of the parallel linear transformation BPLF is as shown in figure 2.

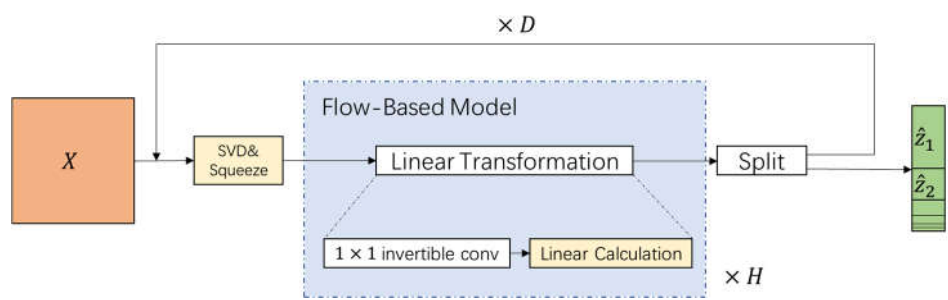

**Figure 2.** The overall structure of BPLF. The diagram shows the process from sample to latent variable. X:     SVD:     D:     H:     z1:     z2:     (annotation)

According to the research byGlow, in order to ensure that the information between different channels can be merged with each other, 1*1 reversible convolution is necessary.    Thereforeit is also retained in BPLF.

In many deep generative models, the specific direction of generation is controlled according to  priori conditions. Such requirements have more practical applicable scenarios. In the various generative models introduced before, there are corresponding conditional versions, such as Conditional GAN[16], Conditional VAE[13], Conditional PixCNN[30]. The main idea is to introduce the conditional variable h in the training process. Variable h can be used in many diferent ways including One-hot encoding, the category label, part of the image data (which can be used for image restoration), and data of different modalities. This paper proposes to use the condition variable h as the category label, which can be regarded as an improved version of purely unsupervised BPLF. Given the prior condition h, its linear transformation is as below:

$$y_k = s(x_{k-1}, h) \odot x_k + \mu(x_{k-1}, h) \qquad (3\text{-}8)$$

It can be seen that the parameters s and μ are all added with the prior condition and h is used as additional input. From this formula, it can be seen that the inverse calculation based on the linear transformation is easy to calculate.

## 4. Experiment and data analysis

### 4.1 Dataset

The generative model requires a lot of data for training. In order to prove that the generative model proposed in this paper is applicable, first set of experiments are carried out on the published data image set including MNIST, CIFAR-10. Then it is expanded on the basis of the public facial emotion image data set, and integrated into the emotion image data set. EmotionSet is needed to train the generative model to carry out the facial emotion image generation. Our experiments have included Jaffe[31], CK+[32], Fer2013[33], MMI[34].

The Jaffe dataset is a facial emotion image dataset established by Lyons et al. It includes a total of 213 facial expressions of 10 Japanese female students. The images are mainly obtained in a laboratory environment. Researchers give certain stimuli to the experimenters and capture the original image in frontal shooting. This data set has certain problems. For example, in the experiment, the gray level of each image is inconsistent due to lighting reasons. The experiment only analyzed images of young Japanese women, and the facial emotions of other ethnic groups are lacking. Thus, it is theoretically unfeasible to use only one of the data sets to generate facial emotion images.

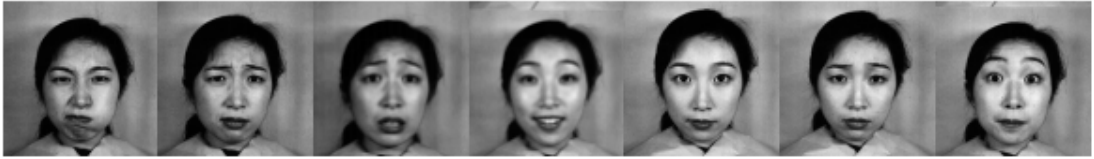

Figure 3 Examples of seven expressions of the same face in Jaffe image set

The CK+ (The Extended Cohn-Kanade) data set is collected by P.Lucy. Unlike Jaffe, the CK+ image data set is mainly extracted from the videos of 118 different experimenters, of which the male accounted for 31% and the female accounted for 69%. CK+ is considered to be the most widely used facial emotion classification data set generated in the laboratory, and is currently used in most facial emotion recognition methods.

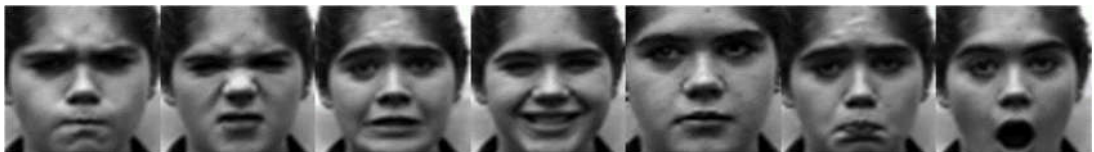

Figure 4 Examples of seven expressions of the same face in CK+

FER (Facial Expression Recognition) 2013 is a dataset used in the 2013 Kaggle Facial Emotion Recognition Challenge.

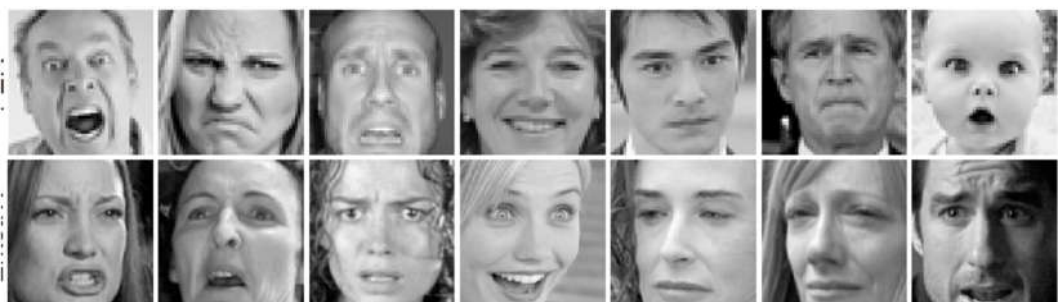

Figure 5 Examples of seven expressions in Fer2013

The MMI-facial expression database (MMI) contains more than 2,900 videos under 75 topics, which are encoded at the frame level, indicating whether each frame is in the neutral, starting, vertex, or offset stage. Ten-fold cross-validation that is not relevant with identity was originally used in this data set, andis merged in this paper.

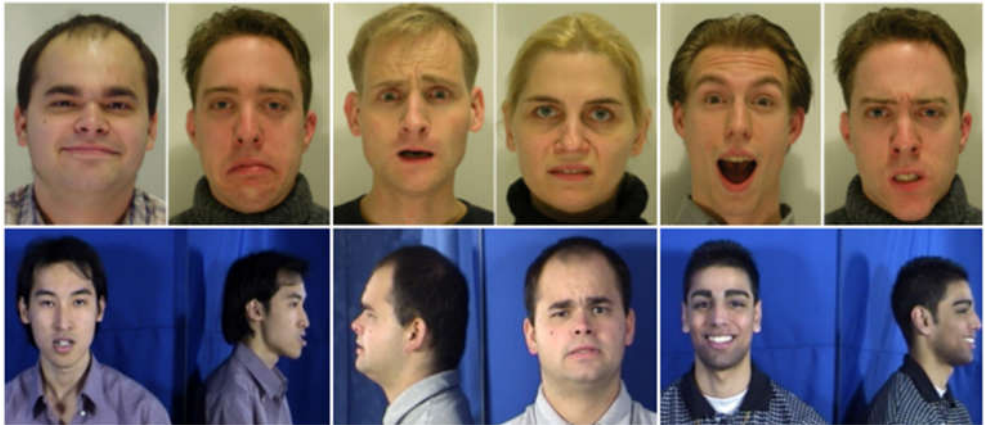

Figure 6 Examples of six expressions in MMI

Since the current published face emotion data set is mainly used for classification training, it is inevitable that there will be some noise interference. The pros and cons of the data set are inseparable from the quality of the generated model. In order to improve the quality of the data set for training, our research rearranges the 4 types of public face emotion data sets. Table 1 shows the statistical information about the number of samples in the original data set.

Table 1 Statistics on the sample size of the four data sets

| Label\Dataset | Jaffe | CK+ | Fer2013 | MMI |
|---|---|---|---|---|
| Anger | 30 | 136 | 3927,354,654 | 96 |
| Disgust | 29 | 178 | 436,55,55 | 144 |
| Fear | 32 | 76 | 4022,355,527 | 96 |
| Happy | 31 | 208 | 7002,895,879 | 224 |
| Neutral | 30 | 174 | 4895,398,626 | 0 |
| Sad | 31 | 85 | 4729,454,594 | 144 |
| Surprise | 30 | 250 | 3129,321,416 | 144 |

FER2013 divides the image data into three parts: training set, validation set and test set. After the statistics are completed, the data set is correspondingly preprocessed, and the basic process is shown in figure 7.

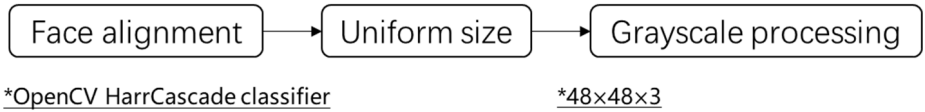
<p style="text-align:center">*OpenCV HarrCascade classifier            *48×48×3</p>

Figure 7 Data set preprocessing process

After the face image preprocessing process, the data scale of the Emotion set dataset used in the training of face emotion images in this article is shown in table 2.

Table 2 Emotion Set sample data statistics

| Label | # of sample |
|---|---|
| Anger | 2334 |
| Disgust | 510 |
| Fear | 2315 |
| Happy | 4751 |
| Neutral | 3660 |
| Sad | 2556 |
| Surprise | 2609 |

## 4.2 Experimental results

When evaluating generative models, there are many evaluation parameters, such as Inception Score, Wasserstein distance, etc. Maximum likelihood and density estimation are often used in flow-based generative models. The process of training the model is to reduce the logarithmic maximum likelihood value, by converting its value to bit/dim to measure the fitting ability of the built model to the probability distribution of the original data set:

$$\frac{bit}{dim} = -(\frac{loglikelihood}{height \times weight \times channel} - \ln 128)/\ln 2 \tag{4-1}$$

The measurement method is density estimation, which is a relative measurement method, where 128 is the normalization of the image, the pixel range of [0, 256] is transformed into [-1, 1], and the actual measurement is 1/128 of the original pixel range. This evaluation method is used as the main evaluation index in NICE, RealNVP, and Glow, and will continue to be used in the experiments of this paper.

Our experiments were carried out using dimensionality reduction method and optimization of the model. In addition, we have made improvements including introducing principal component analysis and proposing a parallel linear flow model . The comparative experiments were carried out. When epoch reaches 2000, the model tends to converge. The experimental design and results are shown in table 3.

Table 3 Results of ablation experiments

| Method\Dataset | MNIST | CIFAR-10 | Emotion set |
|---|---|---|---|
| Glow | 1.18 | 3.53 | 2.24 |
| Glow with SVD | 1.16 | 3.50 | 2.21 |
| BPLF 3*3 conv | 1.14 | 3.42 | 2.16 |
| BPLF 3*3&1*1 conv | 1.15 | 3.40 | 2.14 |
| BPLF 3*3 conv with SVD | 1.13 | 3.33 | 2.12 |
| BPLF 3*3&1*1 conv with SVD | 1.14 | 3.38 | 2.15 |

SVD means that singular value decomposition is introduced in the transformation. 3*3 conv means that three 3*3 convolution kernels are used in each "g". 3*3&1*1 conv means that one 1*1 convolution kernel and two 3*3 convolutions are used. The introduction of 1*1 convolution is not for dimensionality reduction, but to reduce the parameters as many as possible without affecting the expressive ability of the network. The experimental results are shown in Figure 4.9. After introducing singular value decomposition, only 3*3 convolution has achieved the best effect in this experiment. It can be seen from the data that the number of network layers is not deep. Using more convolution kernels is conducive to extract richer feature information. The figure below shows that the NLL loss change in 1500-2000 epoch during the training process with Emotion set as the training set. The smooth is 0.85.

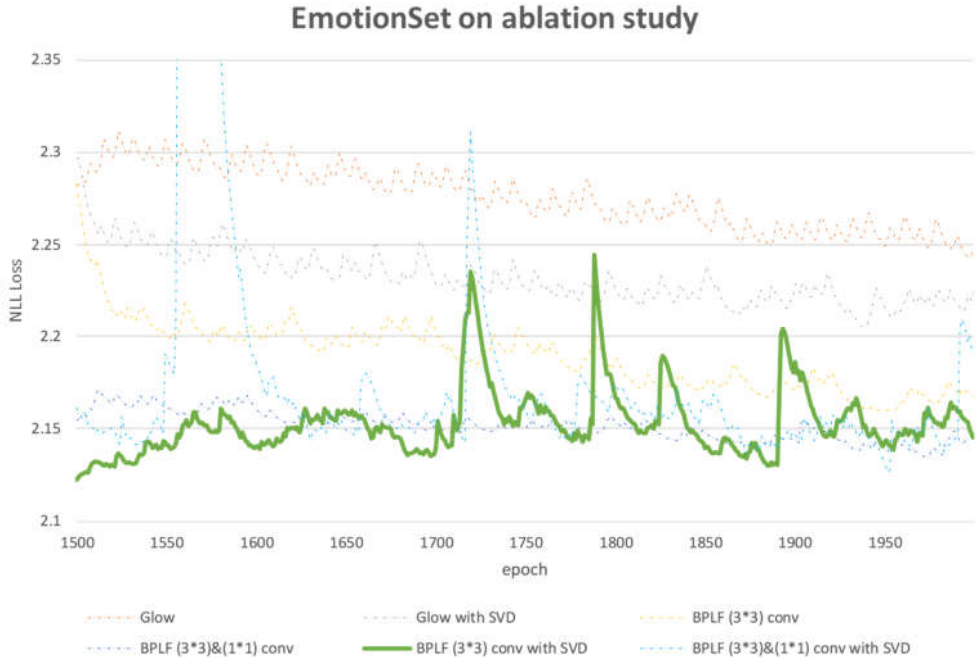

Figure 7 The training process of different models in the ablation experiment

It can be seen from the figure that in the same epoch, BPLF (3*3) conv combined with SVD can achieve the best results. Overall, BPLF (3*3) conv with SVD also achieves the best performance. On this basis, we used this model to compare the two quantitative evaluation indicators of log likelihood and density evaluation with other flow-based generative models. Compared with all data set preprocessing methods, this article is consistent with the Glow paper, as shown in the tables below. BPLF is implemented as a full 3*3 convolution with SVD.

Table 4 Comparison of log-likelihood results among flow-based models. The experimental dataset uses 8-bit encoding

| Model | Glow | f-VAEs | DLF[35] | Flow++[36] | BPLF |
| --- | --- | --- | --- | --- | --- |
| MNIST | 3215.3 | 3116.4 | 3134.6 | 3473.6 | 3124.6 |
| CIFAR-10 | 4513.2 | 4376.9 | 4104.5 | 4492.3 | 4212.4 |
| EmotionSet | 4618.2 | 4515.4 | 4500.7 | 4592.4 | 4498.3 |

Table 5 Performance comparison of density evaluation (bits/dim) among flow-based models. The experimental data set uses 8-bit encoding

| Model | Glow | f-VAEs | DLF | Flow++ | BPLF |
| --- | --- | --- | --- | --- | --- |
| MNIST | 1.24 | 1.15 | 1.14 | 1.14 | 1.13 |
| CIFAR-10 | 3.35 | 3.34 | 3.11 | 3.26 | 3.26 |
| EmotionSet | 2.24 | 2.14 | 2.13 | 2.14 | 2.12 |

In these three types of datasets, it can be found that BPLF is easier to achieve good results in the grayscale dataset. In the environment where the GPU is RXT 2080 Ti, when the model is trained to the maximum round epoch=2000, it takes 3 Weeks, when the model tends to converge. Compared with the Glow model (both epoch=1000), the biggest improvement of BPLF is to reduce the noise of the image, which makes the image look smoother. Figure 8 shows that a comparison of samples randomly generated by the two models.

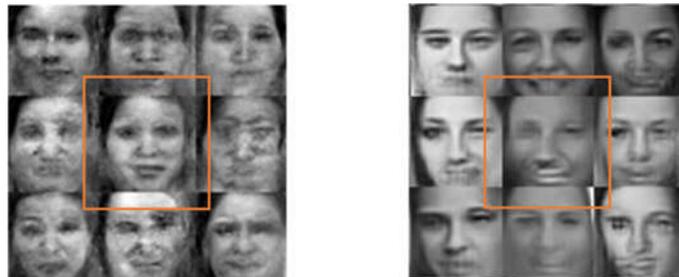

Figure 8 Emotion set has sample images randomly generated with label information. Left image is Glow, and the right image is BPLF

This article uses the EmotionSet data set of category labels to conduct experiments. As shown in the figures below, the one-hot encoded information is used as the category label. The corresponding labels are: anger, disgust, fear, happiness, sadness, surprise. Among the seven emotions, happiness, neutrality and anger showed better results. Some facial features even have corresponding details. And the light and shadow of the face is softer than other models. Since the generation effect of the flow-based generative model is very dependent on the training data, the generation effect of disgust is slightly weaker than other classifications. It is found that, in the EmotionSet data set, the network first recognizes the initial face as a neutral face. The subsequent sampling is the conversion to different labels . This can be regarded as the model's learning and understanding of data features, consistent with human perception.

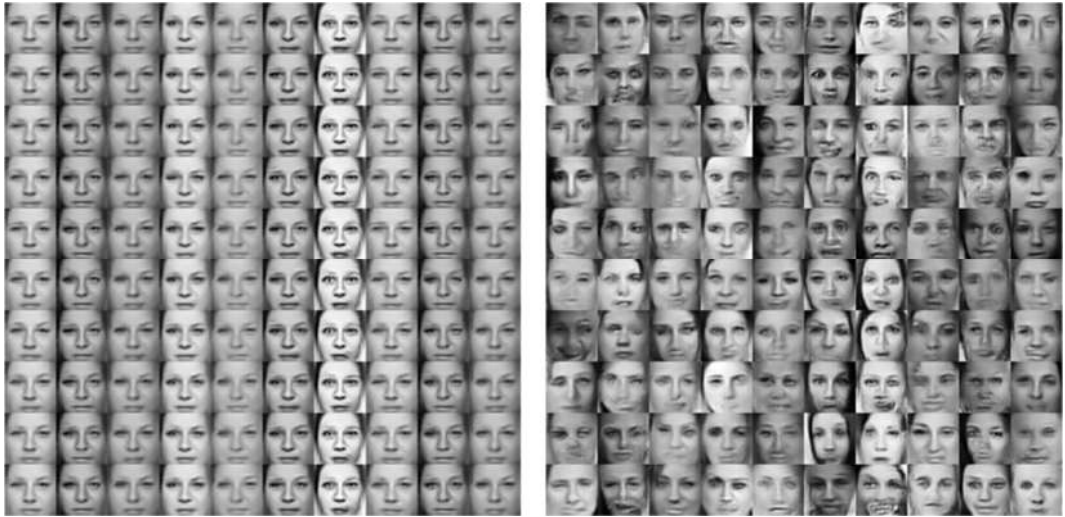
Figure 9 Randomly generated sample images of unlabeled BPLF on EmotionSet

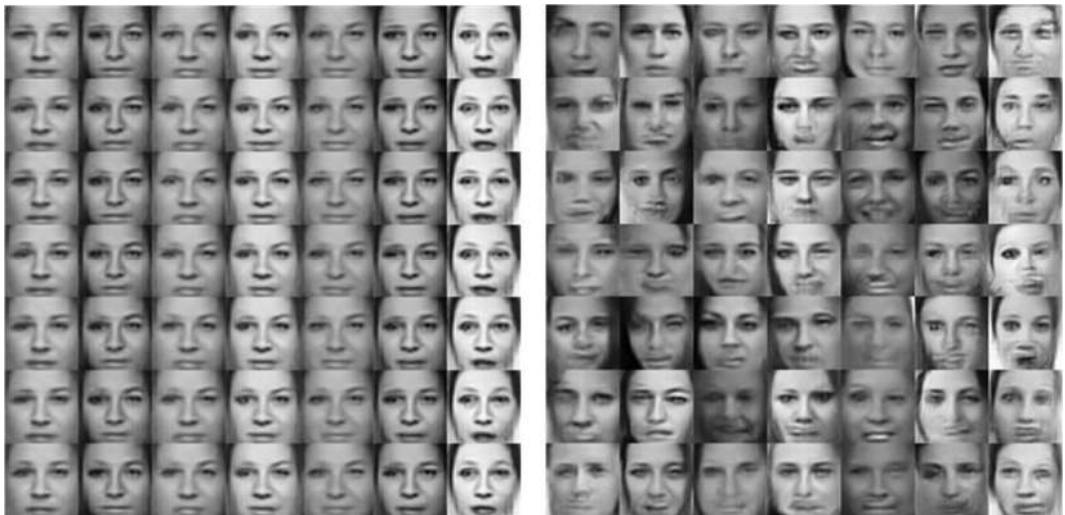
Figure 10 Randomly generated sample images of BPLF with labels on EmotionSet

## 5. Conclusions

In the field of facial emotion research, most of them are still based on recognition. However, if you want to truly study the internal connection of the data, it is necessary to model the data from different angles. Based on this idea, we constructed a deep model. Our work has proposed the idea of BPLF and improved the flow-based generative model. Compared with the previous generative model, the identity transformation is removed and the model capability is enhanced. Linear expressionensures the parallel computing capabilities of the model.  On this basis, the introduction of principal component analysis in the model training process accelerates the convergence of the model. To accomplish these improvements, a series of comparative experiments are designed in our research. Our experimental results showed that the best results are achieved by using 3 layers of 3*3 convolution kernels in BPLF.

From the generated samples, it can be seen that the flow-based generation model can generate more natural images, which is superior compared with the current mainstream depth generation model. Our work not only provides additional possibility for generative model modeling, but also proposes new research ideas for facial emotion research.。